\documentclass{ectj}

\usepackage{amsfonts,amssymb,graphics,epsfig,verbatim,bm,latexsym,amsmath,url,amsbsy}

\usepackage{empheq}
\usepackage{bbm}

\usepackage{lastpage}
\usepackage{wrapfig}
\usepackage{cancel}
\usepackage[justification=centering]{caption}
\usepackage[retainorgcmds]{IEEEtrantools}
\usepackage[margin=3cm]{geometry}
\usepackage{pdfpages}
\usepackage{transparent}

\renewcommand{\thesection}{\arabic{section}}
\renewcommand{\theequation}{\arabic{section}.\arabic{equation}}

\received{...}
\accepted{...}
\volume{21}

\title[Discrete Choice \& Machine Learning]{Discrete Choice Analysis with Machine Learning Capabilities}

\author[Aboutaleb et al.]{Youssef M. Aboutaleb$^{\dagger}$, Mazen Danaf$^{\dagger}$, Yifei Xie$^{\dagger}$, Moshe E. Ben-Akiva $^{\dagger}$}

\address{$^{\dagger}$ Massachusetts Institute of Technology,
            77 Massachusetts Avenue, Cambridge, MA 02139, USA.}
\email{ymedhat@mit.edu, mdanaf@mit.edu, yifeix@mit.edu, mba@mit.edu}

\def\AmSTeX{$\cal A$\kern-.1667em\lower.5ex\hbox{$\cal M$}\kern-.125em
    $\cal S$-\TeX}
\def\BibTeX{{\rm B\kern-.05em{\sc i\kern-.025em b}\kern-.08em
    T\kern-.1667em\lower.7ex\hbox{\mathbf{E}}\kern-.125emX}}
\begin{document}
    \begin{abstract}
This paper discusses capabilities that are essential to models applied in policy analysis settings and the limitations of \textit{direct} applications of off-the-shelf machine learning methodologies to such settings. Traditional econometric methodologies for building discrete choice models for policy analysis involve combining data with modeling assumptions guided by subject-matter considerations. Such considerations are typically most useful in specifying the \textit{systematic} component of random utility discrete choice models but are typically of limited aid in determining the form of the \textit{random} component.  We identify an area where machine learning paradigms can be leveraged, namely in specifying and systematically selecting the best  specification of the \textit{random }component of the utility equations. We review two recent novel applications where mixed-integer optimization and cross-validation are used to algorithmically select optimal specifications for the random utility components of nested logit and logit mixture models subject to interpretability constraints. 
\keywords{Discrete Choice, Machine Learning, Policy Analysis, Algorithmic Model Selection}

    \end{abstract}


\section{Introduction}

 Machine learning techniques are increasing our capacity to discover complex nonlinear patterns in high-dimensional data; see \cite{bishop2006pattern} and \cite{hastie2009elements}. The impressive predictive powers of machine learning have found useful applications in many fields. It is natural to reflect on whether and how these techniques can be applied to advance the field of discrete choice analysis. 

Traditional machine learning techniques are built for prediction problems. Prediction, an \textit{associational} (or correlational) concept, can be addressed through sophisticated data fitting techniques (\citealp{pearl2000causality}). 
Discrete choice models, on the other hand, are typically deployed in policy analysis settings (\citealp{manski2013public}). Policy analysis demands answers to questions that can only be resolved by establishing a sense of \textit{causation}. To draw conclusions requires that data be combined with sufficient domain knowledge assumptions.
 
Algorithms of systematic data-driven model selection and ideas of cross-validation and regularization are prominent in machine learning methodologies. The appeal of such notions and methods over the sometimes arbitrary specification decisions in traditional econometric models remains; see \cite{athey2018impact}.

The goal of this paper is two-fold. The first is to clearly lay out the main capabilities required of (discrete choice) models developed for policy analysis and demonstrate some of the inadequacies of direct applications of off-the-shelf machine learning techniques to such settings. The second goal is to describe a framework where machine learning capabilities can be used to enhance the predictive powers of traditional discrete choice models without compromising their interpretability or suitability for policy analysis. We present two applications of this approach 
namely in automating the specification of the random component of the utility equations in nested logit (\citealp{aboutalebmsthesis}) and mixed logit (\citealp{aboutalebsparse}). 

\paragraph{Organization of this paper}
 \begin{itemize}
    \item \textbf{Section 2} introduces three levels of questions of interest in a typical policy analysis setting. A primer on supervised machine learning is presented along with a reflection on the core methodological differences between theory-driven econometric models such as discrete choice models and the data-fitting methodologies of machine learning.
    \item \textbf{Section 3} presents, in detail, typical capabilities required of models used for policy analysis and demonstrates the inadequacy of off-the-shelf supervised machine learning. 
    \item \textbf{Section 4} reviews recent attempts in the literature to apply machine learning techniques to discrete choice analysis.
    \item \textbf{Section 5} identifies appealing capabilities of machine learning and presents the incorporation of such capabilities to the nested logit and logit mixture models.
    \item \textbf{Section 6} summarizes the main conclusions and take-aways of this paper.
\end{itemize}
\section{Background}
\paragraph{The inference problem} Consider a population of interest whose members are characterized by features (covariates) $\textbf{x}$ in an input space $\mathcal{X}$ and outcome (response) $y$ in an output space $\mathcal{Y}$ with some joint probability distribution $\mathbb{P}(\textbf{x}, y)$ which is assumed to exist but is not necessarily known a priori. 
The classical inference problem of interest is to infer outcome $y$ as a function of features $\textbf{x}$. This generally entails learning (some function of) the conditional probability distribution $\mathbb{P}(y|\mathbf{x})$.

The conditional distribution provides the researcher of a model of the population under study. Three questions could be asked of this model: 
\begin{itemize}
    \item[Q1] What is the distribution of $y$ conditional on some \textit{observed} value of $\textbf{x}_{obs}$?
    \item[Q2]What is the distribution of $y$ conditional on an \textit{extrapolated} value $\textbf{x}_{ext}$ off the support of $\mathbb{P}(\textbf{x})$?
    \item[Q3]What is the distribution of $y$ given an \textit{intervention} that sets the value of $\textbf{x}$ to $\textbf{x}_{int}$?
    
\end{itemize}
It will be clear through this paper that off-the-shelf supervised machine learning, unguided by theory, can only reliably address the first question-- which is a prediction question. While policy analysis applications are typically also concerned with the second and third questions. 

\paragraph{Supervised Machine Learning}
The paradigm of supervised machine learning is that of \textit{learning to predict by example}.  Given an i.i.d sample of input/output pairs $\mathcal{D}=\{(\textbf{x}_i,y_i)\}_{i=1}^n$, called training data, the problem of supervised learning is that of finding a well-fitting function $\hat{f}:\mathcal{X} \rightarrow \mathcal{Y}$. The fitted function $\hat{f}$ is said to generalize well if $\hat{f}(\textbf{x})$ is a good estimate of $y$ on data pairs $(\textbf{x},y)$ drawn according to $\mathbb{P}(\textbf{x}, y)$ and not limited to the specific pairs in the training sample $\mathcal{D}$. 

This paradigm requires specifying a loss function $\ell: \mathcal{Y}\times\mathcal{Y}: \rightarrow [0,\infty)$ for measuring the quality of predictions. This provides an objective measure for choosing $f$. The risk of a function $f$ is defined as the expected loss over the distribution of values the data pairs can take:
\begin{equation}
    {R}(f)=\int_{\mathcal{X}\times \mathcal{Y}}\ell(y,f(\textbf{x}))\mathbb{P}(d\textbf{x},dy)
\end{equation}
The squared loss $\ell(y,f(\textbf{x}))=(y-f(\textbf{x}))^2$ is typical for prediction tasks, where $\mathcal{Y}=\mathbb{R}$, and the logistic loss $\ell(y,f(\textbf{x}))=log(2+exp(-yf(\textbf{x})))$ is typical for classification tasks, where $\mathcal{Y}= \{-1,1\}$.

The problem of supervised learning is then to solve:
\begin{equation}
    \min_{f} R(f)
\end{equation}
given only $\mathcal{D}$. An exact solution for general functions $f$ and losses $\ell$ is clearly not possible. Another complication is that the joint distribution over which the expectation is taken is not known a priori. A path for tractability then is to restrict functions $f$ to some hypothesis space $\mathcal{H}$, for example the linear functions $f(x)=\beta^T\textbf{x}$, and to replace the expected risk by the \textit{empirical} risk calculated from the data:
\begin{equation}
    \hat{R}(f)=\frac{1}{n}\sum_{i=1}^n\ell(y_i,f(\textbf{x}_i))
\end{equation}
The learning problem is then approximated by minimizing the empirical risk over a restricted hypothesis space $\mathcal{H}$:
\begin{equation}
    \min_{f\subset \mathcal{H}}  \hat{R}(f)
\end{equation}
Since the sampled data $\mathcal{D}$ are random and in practice the measurement pairs are noisy, if the hypothesis space is large relative to the sample size, it can happen that the empirical risk is not a good approximation to the expected risk. A typical behavior is that $f$ fits to noise in the observed sample and
\begin{equation}
    \min_{f\subset \mathcal{H}}  \hat{R}(f) \ll \min_{f} R(f)
\end{equation}
This phenomenon known as over-fitting. A way of mitigating against this is to consider a regularizer $G:\mathcal{H}:\rightarrow [0,\infty)$ that penalizes complexity in $f$. The objective function in (4) is replaced by 
\begin{equation}
    \min_{f\subset \mathcal{H}}  \hat{R}(f) +\lambda G(f)
\end{equation}
for $\lambda >0$ \cite{MIT9520}.\\The parameter $\lambda$ is determined through a procedure known as \textit{cross-validation}. The main idea is that since the empirical risk evaluated on the training data (training loss) is not a good approximation to the true risk, a random sample is held out from $\mathcal{D}$ and used to approximate the true risk at solutions to (6) for various values of $\lambda$. The evaluation of the empirical risk (3) approximation on this independent hold out sample is called the validation loss. The optimal amount of regularization $\lambda$ is chosen so that the validation loss is minimized. Such $\lambda$ balances the complexity and `generalizability' of the function $f$ for the given learning task. There is a trade-off: \textit{overly} complex functions $f$ tend to fit to noise and generalize poorly. The \textit{right} amount of penalty applied to the complexity gives a best fitting generalizable model \cite{hasti2001elements}. Cross-validation is a method to determine this right amount of penalty. There are other approaches to prevent overfitting, these include model averaging (`boosting') and estimating separate models on subsamples of the data (`bagging') \cite{athey2018impact}.

The model does not need to learn the entire conditional distribution to make a prediction. The conditional mean or quantile is usually sufficient depending on the choice of the loss function $\ell$. Indeed the solution to (2) for the squared loss error can be shown to be simply the conditional expectation over the conditional distribution \cite{hasti2001elements}:
\begin{equation}
    f(\textbf{x})= \mathbb{E}(y|\mathbf{x})
\end{equation}
This is also known as the regression function.

 Depending on the use-case and the sample size, the hypothesis space $\mathcal{H}$ can be adapted to accommodate general functions with severe non-linearities. Two of the common possibilities include:
\begin{enumerate}
    \item $f(\textbf{x})=\beta^T\phi(\textbf{x})$
    \item $f(\textbf{x})=\phi(\beta^T\textbf{x})$
\end{enumerate}
Where $\phi(.)$ is a non-linear function. Noting that the latter choice can be iterated $f(\textbf{x})=\phi(\beta_L^T\phi(\beta_{L-1}^T\ldots \phi(\beta_1^T\textbf{x})))$ to arrive at a basic multi-layer neural net \cite{MIT9520}.

In addition to the choice of hypothesis space $\mathcal{H}$, there are two main modeling assumptions:
\begin{enumerate}
    \item The data are drawn independently.
    \item The data are identically distributed- there exists a \textit{fixed} underlying distribution.
\end{enumerate}

The appeal of supervised machine learning is in its ability to perform well on prediction tasks by fitting complicated and generalizable functional forms to discover sophisticated patterns in the data with little specification or input from the user.  The success of supervised machine learning, however, hinges on some form of biased estimation. The bias is a direct result of regularization which trades off parameter un-biasedness for lower prediction variance \cite{breiman2001statistical}.

The \textit{i.i.d} assumption holds so long as prediction is limited to features drawn according to the same fixed joint distribution that generated the data used in the training procedure. Supervised machine learning models are therefore excellent candidates for answering questions of the type:
\begin{quote}
    Q1 What is the distribution of $y$ conditional on some observed value of $\textbf{x}_{obs}$?
\end{quote}
absent any interpretability considerations.
\paragraph{Discrete choice models and the econometric approach}
Discrete choice models deal with inference problems where the output space is discrete or categorical $\mathcal{Y}=\{1,2,3,...,T\}$ for $T\in\mathbb{N}$. The researcher observes choices made by a population of decision makers. Under the widely adopted random utility maximization framework \cite{mcfadden-1981}, each decision maker ranks the alternatives in the choice set in order of preference as represented by a utility function. Each alternative is characterized by a utility and is chosen if and only if its utility exceeds the utility of all other alternatives \cite{ben1985discrete}. 
Each utility equation includes a random error term, because it is not possible to model every aspect of an alternative or the decision maker in the utility equation.

 In contrast to the supervised machine learning approach, the traditional econometric approach to the inference problem is a more theory-driven process. This involves building a structural model for $\mathbb{P}(y|\mathbf{x})$--combining data with subject-matter assumptions and knowledge of the sampling process through which the data was obtained. These assumptions guide the specification of the \textit{systematic} component of the utility equations and the handling of potential selection bias or endogeneity. Under this paradigm, transparent models with a strong theoretical base are the ideal. 

 It is understood that there might well be present countless influences, non-linearities, missing attributes and heterogeneities that are unaccounted for in the systematic part. A stochastic or random component will also need to be incorporated to account for such aspects. A few alternative model specifications are estimated on the full dataset and statistical theory is used to determine goodness of fit. A main consideration of model specification and estimation is the recovery of unbiased, or at least consistent, estimates of the policy parameters of interest.

The parameters of the estimated models carry clear subject-matter interpretations. These are subjected to sanity checks (the signs and relative magnitudes for example) and a determination is made as to whether the systematic or random specifications need to be modified. Often, a number of revisions are required before the model is deemed fit-for-use from a policy analysis perspective.

In essence, econometric model building is an effort to create \textit{causal} models \cite{angrist2008mostly} \cite{greene2003econometric}. With the understanding that identifying causality from observational data is at best somewhat tentative and must be combined with assumptions founded on subject-matter assumptions and knowledge of the sampling process \cite{manski2009identification}. From this stand point, empirical fit is not the only consideration for model choice.


\section{Models for Analysis}

Discrete choice policy analysis aims to predict behaviour in counterfactual settings where the attributes of alternatives or the characteristics of current decision makers change, new alternatives become available, existing ones become unavailable, or new decision makers arise \cite{manski2013public}. Policy analysis settings present hypothetical what-if questions such as ``\textit{what will happen if we raise transit fares?}". The answers to such questions requires models that can infer consequences of actions, i.e. models that capture a sense of causation. Different causal mechanisms have different implications for policy. 

We motivate this discussion by quoting an example from \cite{manski2009identification}:
\begin{quote}
\textit{Suppose that you observe the almost simultaneous movements of a person and of his image in a mirror. Does the mirror image induce the person's movements, does the image reflect the person's movements, or do the person and image move together in response to a common external stimulus? Data alone can not answer this question. Even if you were able to observe innumerable instances in which persons and their mirror images move together, you would not be able to logically deduce the process at work. To reach a conclusion requires that you understand something of optics and of human behavior.}
\end{quote}
A model that only captures correlations or associations will rightly predict that an image will always appear whenever a person moves in front of a mirror. However, it can not correctly infer the effect on the person's movements of an intervention--say the shattering of that mirror. ``No image, therefore no motion!'' falsely concludes the correlational model with high confidence. Such is the kind of hypothetical what-if questions of policy analysis (although typically of a more constructive type).

Supervised learning models are only optimized directly to capture correlations: a function $f$ is chosen so that the risk (2) is minimized over the joint distribution.
As the reflection problem shows, addressing what-if extrapolative or interventional questions based on observational data requires that these observations be combined with assumptions on the underlying generating mechanisms \cite{manski1993identification}:
\begin{quote}
    \textbf{Data + Assumptions $\rightarrow{}$ Conclusions}
\end{quote}
The only other resolution being the initiation of a new sampling processes to collect experimental data, which is typically impractical in many policy settings \cite{manski2009identification}.
 
 The next sections discuss features that are essential to models deployed in policy analysis settings. We argue that these models must provide meaningful extrapolations ( Section 3.1), answers to interventions (Section 3.2), and must be interpretable (Section 3.3).  

\subsection{Extrapolation: Theory as a Substitute for Data}

\begin{figure}[h]
\centering
\includegraphics[scale=0.3]{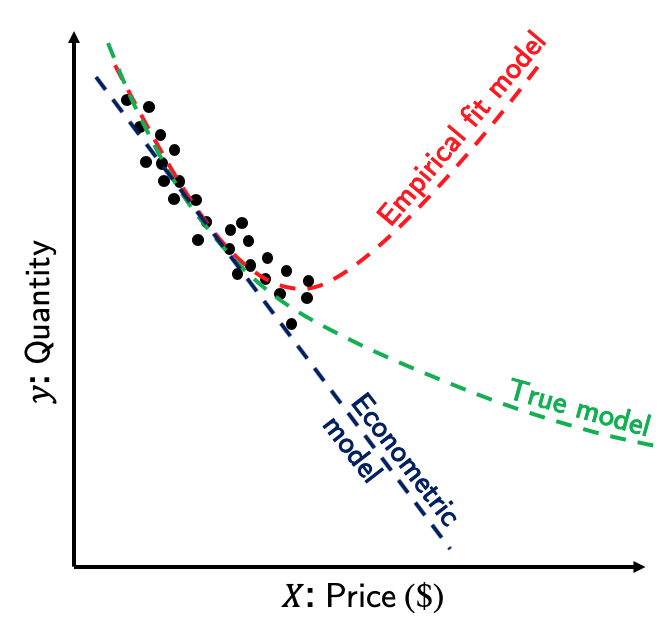}
  \caption{The shape of an empirically fitted model is only governed by the cloud of training data points. Without meaningful restrictions, extrapolations off the training range are hard to justify.}
  \label{fig:extrapolation}
\end{figure}

Consider the demand $y$ of a commodity or service modelled as function of its price $x$ shown in Figure \ref{fig:extrapolation}. The goal is to determine how the demand will respond to changes in price perhaps due to a proposed tax. It is very typical that the range of values over which prices were observed is limited-- prices just do not change enough. The goal is to build a model relating demand to price, a function of $\mathbb{P}(y|x)$, and use this model to extrapolate values of $y$ for values of $x$ outside the range of historically observed prices. 

The supervised machine learning paradigm is one of maximizing fit. A model will be to chosen capture the non-linear trend in the observed data-- perhaps the second order polynomial shown in red in Figure \ref{fig:extrapolation}. This model, chosen to maximize empirical fit, is perfectly suitable for studying how the demand changes for different price points within the locality of historically observed prices. Extrapolations outside that range, without sufficient assumptions, are hard to justify as we will make precise why shortly. 

An econometric approach to this problem will start with a theory-- that demand for a product responds negatively to increases in its price. The negative estimated slope of the simple linear model used, the blue line in Figure \ref{fig:extrapolation}, confirms the researcher's a priori expectations. Extrapolations based off this model are based on a theory which is most needed when making predictions outside the range of observed values. To quote Hal Varian \cite{varian1993use}:
\begin{quote}
    \textit{Naive empiricism can only predict what has happened in the past. It is the theory---the
underlying model---that allows us to extrapolate.}
\end{quote}

Model specifications that maximize fit as the only consideration are not enough to provide meaningful extrapolations. To make this argument more precise consider the general inference setting described in Section 1, and suppose we seek to answer the second question identified:
\begin{quote}
    Q2 What is the distribution of $y$ conditional on an extrapolated value $\textbf{x}_{ext}$ off the support of $\mathbb{P}(\textbf{x})$?
\end{quote}
 The only way to infer $\mathbb{P}(y|\textbf{x}=\textbf{x}_{ext})$ at $\textbf{x}_{ext}$ outside the support of $\mathbb{P}(\textbf{x})$ is to impose assumptions enabling one to deduce $\mathbb{P}(y|\textbf{x}=\textbf{x}_{ext})$ from $\mathbb{P}(y|\textbf{x})$. For concreteness, consider the conditional mean $\mathbb{E}[y|\textbf{x}]$ and look at the two possible ways of its estimation: nonparametric and parametric.

Smoothness regularity assumptions such as continuity or differentiability that enable the \textit{nonparmateric} estimation of $\mathbb{E}[y|\textbf{x}]$ from finite samples imply that $\mathbb{E}[y|\textbf{x}=\textbf{x}_1]$ is near $\mathbb{E}[y|\textbf{x}=\textbf{x}_2]$ when $\textbf{x}_1$ is near $\textbf{x}_2$. This assumption restricts the behaviour of $\mathbb{E}[y|\textbf{x}]$ locally. Let $\textbf{x}_0$ be the point on the support of $\mathbb{P}(\textbf{x})$ nearest to $\textbf{x}_{ext}$. It is not clear whether the distance between $\textbf{x}_0$ and $\textbf{x}_{ext}$ should be interpreted as small enough to be governed by these local restrictions. Extrapolation therefore requires an assumption that restricts the behaviour of $\mathbb{E}[y|\textbf{x}]$ globally. This enables the deduction of $\textbf{x}_{ext}$ from knowledge of $\mathbb{E}[y|\textbf{x}]$ at values of $\textbf{x}$ that are not necessarily near $\textbf{x}_{ext}$ \cite{manski2009identification}.

Recall from Section 1 that a \textit{parametric} estimation of $\mathbb{E}[y|\textbf{x}]$ is obtained by minimizing the squared loss empirical risk over a restricted class of functions $f\subset \mathcal{H}$.
Values of $\textbf{x}$ outside the support of $\mathbb{P}(\textbf{x})$ have no bearing on the value of the empirical risk and therefore have no bearing on the shape of the fitted function outside the support of $\mathbb{P}(\textbf{x})$. In other words, without sufficient restrictions on $\mathcal{H}$, extrapolations off the support are arbitrary. 

Global restrictions make assumptions about how the conditional distribution varies with $\textbf{x}$. These restrictions are chosen by the researcher in line with a priori subject-matter expectations on that relationship. Consider again the conditional mean $\mathbb{E}[y|\textbf{x}]$. The common linear regression assumption is to restrict $\mathbb{E}[y|\textbf{x}]$ to be linear. Other possible assumptions include restricting $\mathbb{E}[y|\textbf{x}]$ to be convex or monotone increasing (in all or some of the covariates $\textbf{x}$).
These and other restrictions enable \textit{meaningful} extrapolations off the support of $\mathbb{P}(\textbf{x})$. 

From this perspective, the primary function of theory is to justify the imposition of global assumptions that enable extrapolation.

\subsection{Intervention: Structural Assumptions Specify Invariant Aspects}

\begin{figure}[h]
\centering
\includegraphics[scale=0.45]{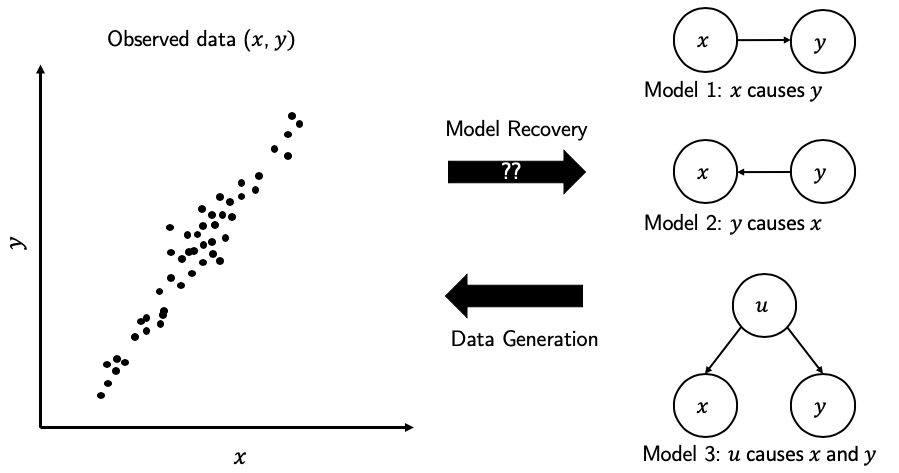}
  \caption{Any number of data generating mechanisms may be consistent with available empirical evidence. The three alternative models on the right produce the same joint distribution of $x$ and $y$. Each model, however, has different implications on how the value of one variable will change in response to an \textit{interventional} policy changing the value of the other variable. This presents an identification problem. Observational data must be combined with structural assumptions, motivated by subject-matter knowledge of $x$ and $y$, for a resolution.}
  \label{fig:identification}
\end{figure}

Suppose variables $x$ and $y$ are observed to be strongly positively correlated as in Figure \ref{fig:identification}. Does $x$ cause $y$? Is it the other way around? or Is there, perhaps, a confounding variable $u$ that causes both $x$ and $y$? Observational data alone can \textit{never} answer this question even if the researcher had access to innumerable observations of pairs $(x,y)$. Yet the underlying data generating process needs to be uncovered before the researcher is able to answer interventional questions. Ignoring this step will lead to misleading conclusions.

An excellent example is provided in \cite{athey2018impact}. Suppose the researcher has access to observational data of hotel room prices and their occupancy rates. Since hotels tend to raise their prices during peak season, occupancy rates are observed to be positively correlated with room prices. Without making any structural assumptions, this data can only answer prediction questions of the first type. For example, an agency seeking to estimate hotel occupancies based on published room rates. What if instead we ask of the model the impact of a proposed luxury tax on occupancy rates? The model will suggest that raising room prices will lead to higher occupancy rates! This an instance of the logical fallacy: \textit{cum hoc ergo propter hoc} (with this, therefore because of this).

What went wrong? Evaluating the effect of interventional policies breaks the assumption of a fixed data generating process that underpins supervised machine learning. Structural assumptions that encode a sense of causality are therefore needed \cite{brockman2019possible}:
\begin{quote}
    \textit{With regard to causal reasoning, we find that you can do very little with any form
of model-blind curve fitting, or any statistical inference, no matter how sophisticated the
fitting process is.}
\end{quote}

Supervised machine learning models, which only learn to capture correlations, can not answer interventional questions which require, in addition to data, strong structural assumptions. Prediction tasks are well managed by these models only under conditions similar to those of the training data $\mathcal{D}$. Recall that one of assumptions of supervised machine learning models is that the data, $\mathcal{D}$, are identically distributed according to some \textit{fixed }joint distribution. The problem of answering interventional questions is that of making predictions under situations that are \textit{changing}-- the assumption that the joint distribution is fixed is not necessarily valid in the ``mutilated" world.  

Answering questions of the third type: \begin{quote}
        Q3 What is the distribution of $y$ given an intervention that sets the value of $\textbf{x}$ to $\textbf{x}_{int}$?
\end{quote}requires combining data with sufficiently strong assumptions on the nature of the modeled world. Nothing in the specification of a joint distribution function $\mathbb{P}(x,y)$ identifies \textit{how} it is to change in response to interventions. This information must be provided by causal assumptions which identify relationships that remain invariant when external conditions change \cite{pearl2000causality}.

\subsection{Interpretability: Amenability to Scrutiny is a Prerequisite to Credibility}

 The ultimate goal of analysis is to uncover insights on the behavior of a population under study--connecting observed data to reality, and to use those insights in forecasting and planning.
 Any model is only a simplification of reality. It will include the salient features of a relationship of interest and will often require a number of sufficient maintained assumptions to meet the demands of policy analysis as discussed in earlier sections. The requirement that the model be used in answering ambitious introspective policy questions sets the bar high. For a model's recommendations to have credibility it must withstand scrutiny. This includes justifications for any assumptions made and an understanding of why the model's output is what it is. 
 
 Trust that the model's results are sensible must first be established before the model is applied to policy analysis. A model's interpretability is its gateway to establishing trust. Interpretable models are amenable to scrutiny-- a prerequisite to credibility. 
 
 Transparent models are the gold standard in interpretability. Transparency entails a full understanding of the model's mechanisms and assumptions. Each of the model's parameters admits intuitive subject-matter explanations. A wrong parameter sign, such as a positive coefficient for cost in a demand model, could be a strong cue that the model may be miss-specified. The researcher knows what is wrong and what to fix. Such an understanding confers a ``certificate of credibility" to the model--a guarantee, in essence, that while the model's predictions may be imprecise, the results are `in the right direction'. With such credibility, trust is established and the model is suitable for policy analysis. 
 
 Black box models are much harder, if not impossible, to fully scrutinize. The parameters of such models are not identifiable and do not carry subject-matter interpretations. It is sometimes still possible to query these models and extract information in a post hoc analysis \cite{lipton2016mythos}. A major problem remains. When the output does not conform to a priori expectations and the results are counter intuitive, the parameters provide no clues as to what went wrong and what should be fixed. It is not clear whether the problem is in training, in method or because things have changed in the environment \cite{pearl2019limitations}.

\section{Direct Machine learning applications to discrete choice}
This section surveys efforts in the literature of applying machine learning paradigms and techniques to models of discrete choice. 
\paragraph{Direct comparisons of fit} Several studies in the literature compare the predictive accuracy of machine learning models such as neural nets and support vector machines to classical discrete choice models (such as flat and nested logit models) in various applications including travel mode choice \cite{zhang2008travel} \cite{omrani2015predicting} \cite{hagenauer2017comparative}, airline itinerary choice \cite{lheritier2019airline}, and car ownership \cite{paredes2017machine}. The unanimous conclusion that machine learning models provide a better fit is hardly a surprise. The usability of these models for policy analysis is suspect as we have demonstrated in the previous section.

\paragraph{Post hoc analysis of black box models} A few studies consider the application of non-transparent models to discrete choice settings and rely on post hoc analysis of output for insight. \cite{van2019using} used a neural network to estimate the value of time distribution using stated choice experiments with a faster/more expensive alternative and a slower/cheaper alternative. The authors claim that this method can uncover the distribution of value of time and its moments without making strong assumptions on the shape of the distribution or the error terms, while incorporating covariates and accounting for panel effects. 

\cite{wang2018using} proposes an empirical method to extract valuable econometric information from neural networks, such as choice probabilities, elasticities, and marginal rates of substitution. Their results show that when economic information is aggregated over the population or ensembled over models, the analysis can reveal roughly S-shaped choice probability curves, and result in a reasonable median value-of-time. 
The authors admit, however, that at the disaggregate level, some of the results are counter-intuitive (such as positive cost and travel time effects on the choice probabilities, and infinite value of time).

\paragraph{Algorithms for big data} A number of researchers studied the use of specific optimization algorithms that are traditionally used to train machine learning models to facilitate the estimation of discrete choice models over large datasets. 

\cite{lederreystochastic} introduced an algorithm called Window Moving Average - Adaptive Batch Size, inspired by Stochastic Gradient Descent, used it to estimate mutlinomial and nested logit models. The improvement in likelihood is evaluated at each step, and the batch size is increased when the improvement is too low using smoothing techniques.

In the context of logit mixture models, \cite{braun2010variational} proposed a variational inference method for estimating models with random coefficients. Variational procedures were developed for empirical Bayes and fully Bayesian inference, by solving a sequence of unconstrained convex optimization problems. After comparing their estimators to those obtained from the standard MCMC - Hierarchical Bayes method \cite{allenby1997introduction} \cite{allenby1998marketing} \cite{train2009discrete} on real and synthetic data, the authors concluded that variational methods achieve accuracy competitive with MCMC at a small fraction of the computational cost. The same conclusions are found by several studies including \cite{bansal2019bayesian},\cite{depraetere2017comparison}, and \cite{tan2017stochastic}. \cite{krueger2019variational} extended this estimator to account for inter- and intra-consumer heterogeneity, however, they noted that the results were noticeably less accurate than those obtained from MCMC, mainly because of the restrictive mean-field assumption of variational Bayes. 

\paragraph{Hybrid machine learning and discrete choice models}
\cite{sifringer2018let} introduced the Learning Multinomial Logit model, where the utility specification of a traditional multinomial logit is augmented with a non-linear representation arising from a neural net. The rationale behind this method was to divide the systematic part of the utility specification into an interpretable part (where the variables are chosen by the modeler), and a black-box part that aims at discovering a good utility specification from available data. This method relies on the fact that mutlinomial logit can be represented as a convolutional neural network with a single layer and linear activation functions. 

\paragraph{Machine learning to inform model specification}\cite{bentz2000neural} showed that a feedforward neural network with softmax output units and shared weights can be viewed as a generalization of the multinomial logit model (MNL). The authors also indicated that the main difference between the two methods lies in the ability of neural nets to model non-linear preferences, without a priori assumptions on the utility function. The authors concluded that the if fitted function is not too complex, it is possible to detect and identify some low order non-linear effects from the neural nets by projecting the function on sub-sets of the input space, and use the results to obtain a better specification for MNL.

\cite{van2019artificial} developed a neural net based approach to investigate decision rule heterogeneity among travelers. The neural nets were trained to recognize the choice patterns of four distinct decision rules: Random Utility Maximization, Random Regret Minimization, Lexicographic, and Random. This method was applied to a Stated Choice experiment on commuters’ value of time, and cross-validation was used to compare the results against those obtained from traditional discrete choice analysis methods. The authors concluded that neural nets can successfully recover decision rule heterogeneity.

\section{Discrete Choice Models with Machine Learning Capabilities}
How can machine learning paradigms be leveraged to advance the field of discrete choice? Our motivation for applications of machine learning to discrete choice is directed both by its limitations-- that without incorporating strong structural assumptions and addressing issues of interpretability, machine learning cannot be used for answering the extrapolative and interventional questions of policy analysis-- and its strengths: machine learning provides \textit{flexibility in model specification}, and\textit{ systematic methods for model selection}.

So far, we have established that:
\begin{enumerate}
    \item  Fully data-driven methodologies need to be tempered with structural assumptions with respect to policy variables of interest.
    \item Imposing meaningful subject-matter global restrictions on the hypothesis space $\mathcal{H}$ allows for meaningful extrapolations.
    \item Structural assumptions are needed to establish causality from observational data 
    \item Stcrutiny, at least with respect to the policy variables, is required to asses the model's fit for use. 
\end{enumerate}
Domain knowledge typically informs such assumptions and restrictions and guides assessments of model suitability. Such knowledge is most applicable in specifying the systematic component of random utility discrete choice models and least applicable in determining the specification of the random component. 
This identifies an area where machine learning paradigms can be leveraged, namely in specifying and systematically selecting the best random utility specification.

The systematic component is specified with a priori expectations on the signs and relative magnitudes of the parameters \cite{ben1985discrete}. For example, addition travel cost and time represent added disutility in travel demand, the parameters corresponding to cost and time are expected to be negative in a linear specification of the model. The value of travel time, calculated as the relative magnitude of these parameters is commonly used to assess model specification. 

\paragraph{Where domain knowledge does not help}

While subject-matter knowledge informs the specification of the systematic utility equations, specifying random aspects of the model can be more challenging. For concreteness, we consider two examples: nested logit and logit mixture models.

First consider the problem of specifying the nesting structure in nested logit models. Researches often use their understanding of the choice situation under study to group `similar' alternatives into nests. Alternatives grouped in the same nest share a common error term accounting for shared similarities not directly included in the systematic component. However, a priori expectations about the optimal nesting structure are sometimes misguided. The correlations in the error components depend largely on the variables entering the systematic part of the utility. If the systematic utility equations account for most of the correlation between two similar alternatives, then grouping these alternatives under the same nest does not necessarily improve over flat logit. The researcher typically tests and report two or three alternative nesting structure specifications for robustness. A comprehensive test of all possible structures is impractical for many modeling situations. 

In logit mixture models, the parameters in the systematic utility equations are treated as random variables-- usually normally distributed with mean and covariance to be estimated from the data. Off-diagonal elements in the covariance matrix indicate that a decision maker's preferences for one attribute are related to their preferences for another attribute \cite{hess2017correlation}. The researcher has some leeway in specifying which of these off-diagonal elements to estimate and which to constraint to zero. In practice, these models are typically estimated under two extreme assumptions: either a full or a diagonal covariance matrix \cite{james2018estimation}. A full co-variance matrix implies correlations between all the distributed parameters, while a diagonal matrix implies that these parameters are uncorrelated.
Ignoring correlations between parameters can distort the estimated distribution of ratios of coefficients, representing the values of willingness-to-pay (WTP) and marginal rates of substitution \cite{hess2017estimation}. In practice, it is usually difficult for researchers to hypothesize which subsets of variables are correlated.

The following sections present machine learning data driven methodologies for algorithmically selecting the random specification of the utility components of nested logit (Section 5.1) and logit mixture models (Section 5.2) subject to interpretability considerations. 
The optimal random specification is determined using optimization techniques, regularization, and out-of-sample validation. The econometric tradition of specifying the systematic component the utility remains. The models remain transparent, and the parameters can be used to estimate trade-offs, willingness to pay values, and elasticities as before. 


\subsection{Learning Structure in Nested Logit Models}

Nested logit is a popular modeling tool in econometrics and transportation science when one wants to model the choice that an individual makes from a set of mutually exclusive alternatives \cite{mcfadden-1981} \cite{ben1985discrete}. The nested logit model provide a flexible modeling structure by allowing for correlations between the random components of the alternatives in the choice set.

In specifying a nested logit model, the researcher hypothesizes a nesting structure over the choice set and proceeds to estimate the model parameters (the coefficients in the utility equations that determine the relative attractiveness of choices to the decision maker). Each nest is associated with a scale parameter (which is also estimated), and quantifies the degree of intra-nest correlation \cite{ben1985discrete}. The nesting structure determines \textit{how} the alternatives are correlated, and the scales determine by \textit{what amount} they are correlated.

The large feasible set of possible nesting structures presents a significant modeling challenge in deciding which nesting structure best reflects the underlying choice behavior of the population. The current \textit{modus operandi} is to use domain knowledge to substantially reduce the feasible set to a small set of candidate structures. This is done at the risk of potentially excluding some ostensibly non-intuitive structures which might actually provide a better description of the choice behaviour of the population under study \cite{koppelman-2006}. This is the core motivation of \cite{aboutalebmsthesis} for taking a more \textit{holistic} view of nested logit model estimation, i.e., one that optimizes over structure as well as parameters.

\cite{aboutalebmsthesis} formulates and solves the nested logit structure learning problem as a mixed-integer nonlinear programming (MINLP) problem-- which entails optimizing not only over the parameters of the model but also over all valid nest structures. In other words, \textit{{rather than assuming a nesting structure a priori, the goal is to reveal this structure from the data}}. To ensure that the learned tree is consistent with utility maximization, the MINLP is constrained so that the scales increase with increasing nesting level. The authors penalize complexity in two ways: the number of nests and the nesting level. The optimal model complexity is chosen through a cross-validation procedure.

In advocating for a data-driven approach for specifying a nested logit structure, we are in no way diminishing the role of the modeler or the importance of domain-specific knowledge in specifying and designing good discrete choice models. Recall that the utility of an alternative to the decision makers under study is given by a sum of a systematic component and a random component. It is the modeler's purview to correctly specify the systematic part of the utility equation. Specifying the random part, however, is a tricky business and the optimal structure may be counter-intuitive. In fact, the optimal error structure is not independent of the specification of the systematic part. If all aspects of the choice behavior that account for correlation between choices can be fully captured in the systematic part, no nesting is needed.
\subsection{Sparse Covariance Estimation in Logit Mixture Models}

Logit mixtures permit the modeling of taste heterogeneity by allowing the model parameters to be randomly distributed across the population under study \cite{train2009discrete}. The modeler's task is to specify the systematic part of the utility equations, as well as the mixing distributions of the distributed parameters and any assumptions on the structure of the covariance matrix.

Researchers typically specify either a full or diagonal covariance matrix. \cite{keane2013comparing} compared different specifications with full, diagonal, and restricted covariance matrices and concluded that a full covariance matrix might not be needed in some cases. They concluded that different specifications fit best on different datasets, which means that  researchers cannot know, without testing, which restrictions to impose.

As the number of combinations of all possible covariance matrix specifications grows super-exponentially with the number of distributed parameters, it is not practically feasible for the modeler to comprehensively compare all possible specifications of the covariance matrix in order to determine an optimal specification to use.

Sparse specifications of the covariance matrix are desirable since the number of covariance elements grows quadratically with the number of distributed parameters. Consequently, sparser models provide efficiency gains in the estimation process compared to estimating a full covariance matrix.

\cite{aboutalebsparse} presents the Mixed-integer Sparse Covariance (MISC) algorithm which uses a mixed-integer program to find an optimal block diagonal covariance matrix structure for any desired sparsity level using Markov Chain Monte Carlo (MCMC) posterior draws from the full covariance matrix. The optimal sparsity level of the covariance matrix is determined using out-of-sample validation. Furthermore, unlike Bayesian Lasso-based penalties in the statistics literature, the method in \cite{aboutalebsparse} does not penalize the non-zero covariances. This is a desirable feature, since penalizing the non-zero covariances may lead to underestimating the heterogeneity in the population under study (the covariance estimates will be biased towards towards zero).

\section{Concluding Remarks}
Supervised machine learning methods emphasize empirical fit as the objective, predictive success being the only criterion as opposed to issues of interpretation or establishing causality. This imposes an intrinsic limitation to the application of such models to policy analysis. Prediction is indeed important from several perspectives. From a policy analysis standpoint, however, the success of a model is best judged from its ability to predict in \textit{new} contexts. We have established the following:

\label{sec:others}
\begin{enumerate}
    \item Machine learning and other empirical models that only maximize fit are excellent candidates for prediction problems where interpretability is not a primary consideration and the prediction is localized to situations directly similar to the training environment. 
    \item Discrete choice models seek to answer ``what-if" extrapolatiove and interventional questions that cannot be fully resolved from observational data alone. Instead data must be combined with domain knowledge assumptions.
    \item Efforts to combine aspects of machine learning with discrete choice methods must not come at the cost of interpretability.
    \item Machine learning concepts such as regularization and cross validation have merit in providing a systematic and principled model selection mechanism.
    \item We presented an implementation of such algorithmic model selection techniques applied to two of the most common discrete choice models: the nested logit and the logit mixture model.
\end{enumerate}

We reviewed recent machine learning inspired methodologies for algorithmically selecting the random specifications in nested logit and logit mixtures that maximize fit subject to interpretability considerations.
The econometric tradition of specifying the systematic component the utility remains. The models remain transparent, and the parameters can be used to estimate trade-offs, willingness to pay values, and elasticities as before. 
We have simply automated what the modeler would ideally like to have done: compare all possible nesting tree (or covariance structure) specifications that `make sense" and choose the best one based on likelihood ratio or some other statistical test. 

    
    \bibliography{references}


\end{document}